\documentclass{llncs}
\usepackage{}
\usepackage{bbding}
\usepackage{amssymb}
\usepackage{amsmath}
\usepackage{color} 
\usepackage{setspace} 
\usepackage{multirow}
\usepackage[]{graphicx}  

\begin{document}
\title{Direct detection of pixel-level myocardial infarction areas via a deep-learning algorithm}
\author{Chenchu Xu\thanks{These authors contributed equally to this work.}\inst{1}, Lei Xu$^\star$\inst{3}, Zhifan Gao\inst{2}, Shen zhao\inst{2}, Heye Zhang\thanks{Corresponding Author: Dr. Heye Zhang (hy.zhang@siat.ac.cn) and Prof. Yanping Zhang (zhangyp2@gmail.com).}\inst{2}, Yanping Zhang$^{\star\star}$\inst{1}, Xiuquan Du\inst{1}, Shu Zhao\inst{1}, Dhanjoo Ghista\inst{4}, Shuo Li\inst{5}}
\institute{$^{1}$ Anhui University, Hefei, China
\\
$^{2}$ Shenzhen Institutes of Advanced Technology,\\ Chinese Academy of Sciences, Shenzhen, China
\\
$^{3}$ Beijing AnZhen Hospital, Beijing, China
\\
$^{4}$ University 2020 Foundation, MA, USA
\\
$^{5}$ University of Western Ontario, London ON, Canada}
\maketitle
\vspace*{-0.2 cm}
\begin{abstract}
Accurate detection of the myocardial infarction (MI) area is crucial for early diagnosis planning and follow-up management. In this study, we propose an end-to-end deep-learning algorithm framework (OF-RNN ) to accurately detect the MI area at the pixel level. Our OF-RNN consists of three different function layers: the heart localization layers, which can accurately and automatically crop the region-of-interest (ROI) sequences, including the left ventricle, using the whole cardiac magnetic resonance image sequences; the motion statistical layers, which are used to build a time-series architecture to capture two types of motion features (at the pixel-level) by integrating the local motion features generated by long short-term memory-recurrent neural networks and the global motion features generated by deep optical flows from the whole ROI sequence, which can effectively characterize myocardial physiologic function; and the fully connected discriminate layers, which use stacked auto-encoders to further learn these features, and they use a softmax classifier to build the correspondences from the motion features to the tissue identities (infarction or not) for each pixel. Through the seamless connection of each layer, our OF-RNN can obtain the area, position, and shape of the MI for each patient. Our proposed framework yielded an overall classification accuracy of 94.35\% at the pixel level, from 114 clinical subjects. These results indicate the potential of our proposed method in aiding standardized MI assessments.
\end{abstract}
\vspace*{-0.4 cm}
\section{Introduction}
\vspace*{-0.1 cm}
There is a great demand for detecting the accurate location of a myocardial ischemia area for better myocardial infarction (MI) diagnosis. The use of magnetic resonance contrast agents based on gadolinium-chelates for visualizing the position and size of scarred myocardium has become `the gold standard' for evaluating the area of the MI \cite{org2002imaging}. However, the contrast agents are not only expensive but also nephrotoxic and neurotoxic and, hence, could damage the health of humans. \cite{wagner2003contrastenhanced}. In routine clinical procedures, and especially for early screening and postoperative assessment, visual assessment is one popular method, but it is subject to high inter-observer variability and is both subjective and non-reproducible. Furthermore, the estimation of the time course of the wall motion remains difficult even for experienced radiologists.
\\Therefore, computer-aided detection systems have been attempted in recent years to automatically analyze the left ventricle (LV) myocardial function quantitatively. This computerized vision can serve to simulate the brain of a trained physician’s intuitive attempts at clinical judgment in a medical setting. Previous MI detection methods have been mainly based on information theoretic measures and Kalman filter approaches \cite{shi2003stochastic}, Bayesian probability model \cite{wang2014direct}, pattern recognition technique \cite{afshin2011assessment}\cite{zhen2015direct}, and biomechanical approaches \cite{wong2015computer}. However, all of these existing methods still fail to directly and accurately identify the position and size of the MI area. More specifically, these methods have not been able to capture sufficient information to establish integrated correspondences between the myocardial motion field and MI area. More recently, unsupervised deep learning feature selection techniques have been successfully used to solve many difficult computer vision problems. The general concept behind deep learning is to learn hierarchical feature representations by first inferring simple representations and then progressively building up more complex representations from the previous level. This method has been successfully applied to the recognition and prediction of prostate cancer, Alzheimer’s disease, and vertebrae and neural foramina stenosis \cite{cai2016multi}.
\\In this study, an end-to-end deep-learning framework has been developed for accurate and direct detection of infarction size at the pixel level using cardiac magnetic resonance (CMR) images. Our method’s contributions and advantages are as follows: (1) for the first time, we propose an MI area detection framework at the pixel level that can give the physician the explicit position, size and shape of the infarcted areas; (2) a feature extraction architecture is used to establish solid correspondences between the myocardial motion field and MI area, which can help in understanding the complex cardiac structure and periodic nature of heart motion; and (3) a unified deep-learning framework can seamlessly fuse different methods and layers to better learn hierarchical feature representations and feature selection. Therefore, our framework has great potential for improving the efficiency of the clinical diagnosis of MI.
\begin{figure}[t!p]
\setlength{\abovecaptionskip}{-0.45 cm}
\setlength{\belowcaptionskip}{-0.55 cm}
\centering
\vspace*{-0.55 cm}
\centerline{\includegraphics[width=1.1\linewidth]{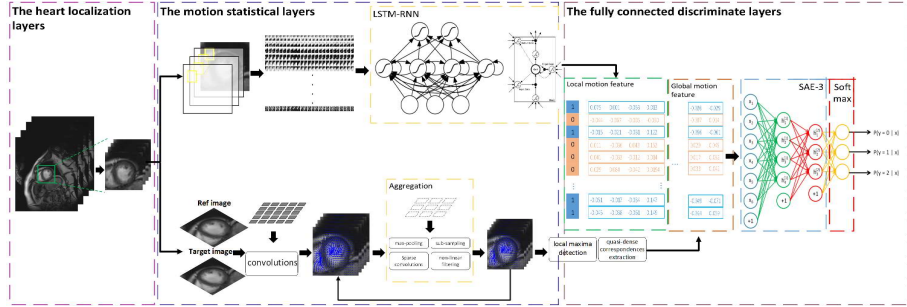}}
\caption{\footnotesize{The architecture of OF-RNN: heart localization layers, motion statistical layers, and fully connected discriminate layers.}}
\label{fig:Figure1}
\vspace*{-0.3 cm}
\end{figure}
\vspace*{-0.3 cm}\section{Methodology}
\vspace*{-0.2 cm}As shown in Fig. 1, there are three function layers inside the OF-RNN. The heart localization layers can automatically detect the ROI, including the LV, and the motion statistical layers can generate motion features that accurately characterize myocardial physiologic and physical function, followed by the fully connected discriminate layers that use stacked auto-encoders and softmax classifiers to detect the MI area from motion features.
\\\textbf{Heart localization layers.} One FAST R-CNN \cite{girshick2015fast} is used here for the automatic detection of a region of interest (ROI) around the LV, to reduce the computational complexity and improve the accuracy. In this study, the first process of the heart localization layers is to generate category-independent region proposals. Afterward, a typical convolutional neural network model is used to produce a convolution feature map by input images. Then, for each object proposed, an ROI pooling layer extracts a fixed-length feature vector from the feature map. The ROI pooling layer uses max pooling to convert the features inside any valid region of interest into a small feature map with a fixed spatial extent of $H$$\times$$W$, where $H$ and $W$ are layer hyper-parameters that are independent of any particular ROI. Finally, each feature vector is fed into a sequence of fully connected layers that branch into two sibling output layers, thereby generating a 64$\times$64 bounding-box for cropping the ROI image sequences, including the LV from CMR sequences.
\\\textbf{Motion statistical layers.}  The motion statistical feature layers are used to extract time-series image motion features through ROI image sequences to understand the periodic nature of ghd heart motion. The local motion features are generated by LSTM-RNN, and the global motion features are generated by deep optical flow.
Thus, in the first step, we attempt to compute the local motion features that are extracted from the ROI image sequence. For each ROI sequence, the input image $I = ({I_1},{I_2}...{I_J},J = 25)$ of size 64$\times$64, $I(p)$ represents a pixel coordinate $p = [x,y]$ of the image $I$. A window of size 11$\times$11 is constructed for the overlapping $I[x,y]$ neighborhoods, which has an intensity value that is representative of the feature of each $p$ on image ${I_J}$. This approach results in the $J$ image sequence features being unrolled as vector ${P_l}(p) \in {R^{11*11*J}}$ for each pixel as input. Then, four layers of RNN \cite{graves2012supervised} with LSTM cells layers are used to learn the input.
Give the input layer ${X_t}$ at time $t$, each time corresponds to each frame(t=J), which indicates that ${x_t} = {P_l}(p)$ at frame $J$, and for the hidden state frame of the previous time step ${h_t-1}$,the hidden and output layers for the current time step are computed as follows:\begin{small}\begin{equation}\setlength{\abovedisplayskip}{2pt}
\setlength{\belowdisplayskip}{2pt}
\begin{array}{l}
{h_t} = \phi \left( {{W_{xh}}\left[ {{h_{t - 1}},{x_t}} \right]} \right),\quad {p_t} = soft\max \left( {{W_{hy}}{h_t}} \right),\quad {\hat y_t} = \arg \max {p_t}\
\end{array}
\end{equation}
\end{small}
where $x_t$, $h_t$ and $y_t$ are layers that represent the input, hidden, and output at each time step $t$, respectively; $W_{xh}$ and $W_{hy}$ are the matrices that denote the weights between the input and hidden layers and between the hidden and output layers, respectively, and $\phi$ denotes the activation function.
The LSTM cell \cite{graves2012supervised} is designed to mitigate the vanishing gradient. In addition to the hidden layer vector $h_t$, the LSTMs maintain a memory vector $c_t$, an input gate $i_t$, a forget gate $f_t$, and an output gate $o_t$; These gates in the LSTMs are computed as follows:\begin{small}
\begin{equation}\setlength{\abovedisplayskip}{2pt}
\setlength{\belowdisplayskip}{2pt}
\left[ \begin{array}{l}
{i_t}\\
{f_t}\\
{o_t}\\
{{\tilde c}_t}
\end{array} \right] = \left( \begin{array}{l}
sigm\\
sigm\\
sigm\\
\tanh
\end{array} \right){W_t}\left[ {D({x_t}),{h_{t - 1}}} \right]
\end{equation}
\end{small}
where ${W_t}$ is the weight matrix, and $D$ is the dropout operator. The final memory cell and the final hidden state are given by\begin{small}\begin{equation}\setlength{\abovedisplayskip}{2pt}
\setlength{\belowdisplayskip}{2pt}
\begin{array}{l}
{c_t} = {f_t} \odot {c_{t - 1}} + {i_t} \odot {\tilde c_t},\quad\quad {h_t} = {o_t} \odot \tanh ({c_t})
\end{array}
\end{equation}
\end{small}
In the second step, we attempt to compute the global motion feature of the image sequence based on an optical flow algorithm \cite{revaud2016deepmatching} by the deep architecture. An optical flow can describe a dense vector field, where a displacement vector is assigned to each pixel, which points to where that pixel can be found in another image. Considering an adjacent frame, a reference image $I$ = ($I_{J-1}$) and a target image $I'$ = (${I_{J}}$), the goal is to estimate the flow $w = {(u,v)^ \top }$ that contains both horizontal and vertical components. We assume that the images are already smoothed by using a Gaussian filter with a standard deviation of $\sigma$. The energy to be optimized is the weighted sum of a data term $ED$, a smoothness term $ES$, and a matching term $EM$:
\begin{small}
\begin{equation}\setlength{\abovedisplayskip}{2pt}
\setlength{\belowdisplayskip}{2pt}
E(w) = \int_\Omega  {{E_D} + \alpha {E_S} + \beta {E_M}dx}
\end{equation}
\end{small}
Next, a procedure is developed to produce a pyramid of response maps, and we start from the optical flow constraint, assuming a constant brightness. A basic way to build a Data term and a Smoothness term is the following:\begin{small}\begin{equation}\setlength{\abovedisplayskip}{2pt}
\setlength{\belowdisplayskip}{2pt}
{E_D} = \delta \Psi \left( {\sum\limits_{i = 1}^c {{w^ \top }} \bar J_0^iw} \right) + \gamma \Psi \left( {\sum\limits_{i = 1}^c {{w^ \top }} \bar J_{xy}^iw} \right)
\end{equation}
\begin{equation}
{E_S} = \Psi (||\nabla u|{|^2} + ||\nabla v|{|^2})
\end{equation}
\end{small}
where $\Psi$ is a robust penalizer; $\bar J_{xy}^iw$ is the tensor for channel $I$; $\delta$ and $\gamma$ are the two balanced weights. The matching term encourages the flow estimation to be similar to a precomputed vector field $w'$, and a term $c(x)$ has been added.\begin{small}\begin{equation}\setlength{\abovedisplayskip}{2pt}
\setlength{\belowdisplayskip}{2pt}
{E_M} = c\Psi ({\left\| {w - {w'}} \right\|^2})
\end{equation}
\end{small}
For any pixel $p'$ of $I'$, ${C_{n,p}}(p')$ is a measure of similarity between $I_{n,p}$ and $I'_{n,p'}$. We have $I_{n,p}$ to be a patch size of N $\times$N(N$\in${4, 8, 16}) from the first image centered at $p$. We start with the bottom-level correlation maps, which are iteratively aggregated to obtain the upper levels. This aggregation consists of max-pooling, sub-sampling, computing a shifted average and non-linear rectification. In the end, for each image $I_{J-1}$, a fully motion field ${{\rm{w}}_{J-1}}{\rm{ =  (}}{{\rm{u}}_{J-1}}{\rm{, }}{{\rm{v}}_{J-1}}{\rm{)}}$ is computed with reference to the next frame ${I_{J }}$.
\\\textbf{Fully connected discriminate layers.} The fully connected discriminate layers are used to detect the MI area accurately from the local motion features and the global motion features. First, for each $w_j$, we use image patches, say 3$\times$3, by extracting the feature beginning from a point $p$ in the first frame and tracing $p$ in the following frame. We can thereby obtain ${P_g}(p)$ while containing a 3$\times$3 vector for displacement and a 3$\times$3 vector for the orientation of $p$ for each frame. Second, we conduct a simple concatenation between the local image feature ${P_l}(p)$ from the LSTM-RNN and the motion trajectories feature ${P_g}(p)$ via optical flow, to establish a whole feature vector $P(p)$. Finally, an auto-encoder with three stacking layers is used for learning the $P(p)$, followed by a softmax layer, which is used to determine whether $p$ belongs to the MI area or not.
\begin{figure}[t!p]
\setlength{\abovecaptionskip}{-0.55 cm}
\setlength{\belowcaptionskip}{-0.55 cm}
\centering
\vspace*{-0.8 cm}
\centerline{\includegraphics[width=0.45\linewidth,height=4.7 cm]{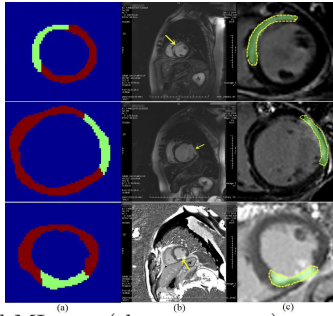}}
\caption{\footnotesize{(a-b) Our predicted MI area (the green zone) can be a good fit for the ground truth (the yellow arrow) (c) our predicted MI area (the green zone) can be a good fit for the ground truth (the yellow dotted line).}}
\label{fig:Figure2}
\vspace*{-0.15 cm}
\end{figure}
\vspace*{-0.2 cm}\section{Experimental results}
\begin{figure}[tp]
\setlength{\abovecaptionskip}{-0.55 cm}
\setlength{\belowcaptionskip}{-0.5 cm}
\centering
\vspace*{-0.6 cm}
\centerline{\includegraphics[width=\linewidth,height=7.8 cm]{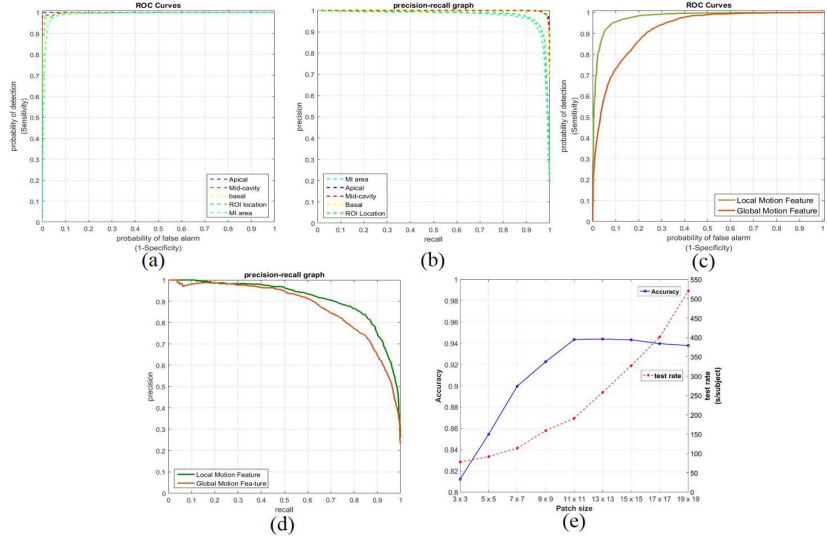}}
\caption{\footnotesize{(a-b) ROCs and PRs show that our results have good classification performance. (c-d) ROCs and PRs for local motion features and global motion features. (e) The accuracy and time for various patch sizes.}}
\label{fig:Figure3}
\vspace*{-0.25 cm}
\end{figure}

\vspace*{-0.15 cm}
\textbf{Data acquisition.} We collected the short axis image dataset and the corresponding enhanced images using gadolinium agents from 114 subjects in this study on a 3T CMR scanner. Each subject’s short-axis image dataset consisted of 25 2D images (a cardiac cycle), a total of 43 apical, 37 mid-cavity and 34 basal short-axis image datasets for 114 subjects. The temporal resolution is 45.1$\pm$8.8 ms, and the short-axis planes are 8-mm thick. The delayed enhancement images were obtained approximately 20 min after intravenous injection of 0.2 mmol/kg gadolinium diethyltriaminepentaacetic acid. A cardiologist (with more than 10 years of experience) analyzed the delayed enhancement images and manually traced the MI area by the pattern of late gadolinium enhancement as the ground truth.
\\\textbf{Implementation details.} We implemented all of the codes using Python and MATLAB R2015b on a Linux (Kylin 14.04) desktop computer with an Intel Xeon CPU E5-2650 and 32 GB DDR2 memory. The graphics card is an NVIDIA Quadro K600, and the deep learning libraries were implemented with Keras (Theano) with RMSProp solver. The training time was 373 minutes, and the testing time was 191 seconds for each subject (25 images).
\\\textbf{Performance evaluation criteria.} We used three types of criteria to measure the performance of the classifier: 1) the receiver operating characteristic (ROC) curve; 2) the precision-recall (PR) curve; 3) for pixel-level accuracy, we assessed the classifier performance with a 10-fold cross-validation test, and for segment-level accuracy, we used 2/3 data for training and the remaining data for testing.
\\\textbf{Automatic localization of the LV.} The experiment's result shows that OF-RNN can obtain good localization of the LV. We achieve an overall classification accuracy of 96.49\%, with a sensitivity of 94.39\% and a specificity of 98.67\%, in locating the LV in the heart localization layers. We used an architecture similar to the Zeiler and Fergus model to pre-train the network. Using selective search’s quality mode, we sweep over 2k proposals per image. Our results for the ROI localization bounding-box from 2.85k CMR images were compared to the ground truth marked by the expert cardiologist. The ROCs and PRs curves are shown in Fig. \ref{fig:Figure3} (a-b).
\\\textbf{MI area detection.} Our approach can also accurately detect the MI area, as shown in Fig. \ref{fig:Figure2}. The overall pixel classification accuracy is 94.35\%, with a sensitivity of 91.23\% and a specificity of 98.42\%. We used the softmax classifier by fine-tuning the motion statistical layers to assess each pixel (as normal/abnormal). We also compared our results to 16 regional myocardial segments (depicted as normal/abnormal) by following the American Heart Association standards. The accuracy performance for the apical slices was an average of 99.2\%; for the mid-cavity slices, it was an average of 98.1\%; and for the basal slices, an average of 97.9\%. The ROCs and PRs of the motion statistical layers are shown in Fig. \ref{fig:Figure3} (a-b).
\\\textbf{Local and global motion statistical features.} A combination of local and global motion statistical features has the potential to improve the results because the features influence one another through a shared representation. To evaluate the effect of motion features, we use local or global motion statistical features separately along with both motion features in our framework. Table \ref{tab:table1} and Fig. \ref{fig:Figure3} (c-d) show that the results that combine motion statistical features in our framework have better accuracy, sensitivity, and specificity in comparison to those that use only the local or global motion features, in another 10-fold cross-validation test.
\begin{table}[hbtp]
\footnotesize
\setlength{\abovecaptionskip}{-0 cm}
\setlength{\belowcaptionskip}{-0 cm}
\vspace*{-0.5 cm}
\centering
  \caption{Combined motion statistical features effectively improve the overall accuracy of our method}
    \begin{tabular}{cccc}
    \hline\noalign{\smallskip}
    local motion feature & $\surd$      &       & $\surd$ \\
    global motion feature &       & $\surd$     & $\surd$ \\
    \hline\noalign{\smallskip}
    accuracy & 92.6\% & 87.3\% & \textbf{94.3\%} \\
    sensitivity & 86.5\% & 79.4\% & \textbf{91.2\%} \\
    specificity & 97.9  & 96.2\% & \textbf{98.4\%} \\
  \hline\noalign{\smallskip}
    \end{tabular}%
  \label{tab:table1}%
  \vspace*{-0.7 cm}
\end{table}%
\\\textbf{Size of patch.} We use an N$\times$N patch to extract the local motion features from the whole image sequence. Because the displacements of the LV wall between two consecutive images are small (approximately 1 or 2 pixels/frame), it is necessary to adjust the size of the patch to capture sufficient local motion information. Fig. \ref{fig:Figure3} (e) shows the accuracy and computational time of our framework, using from 3$\times$3 to 17$\times$17 patches in one 10-fold cross-validation test. We find that the 11$\times$11 patch size in our framework can obtain better accuracy in a reasonable amount of time.
\\\textbf{Performance of the LSTM-RNN.} To evaluate the performance of the LSTM-RNN, we replaced the LSTM-RNN using SVMrbf, SAE-3, DBN-3, CNN and RNN in our deep learning framework, and we ran these different frameworks over 114 subjects using a 10-fold cross-validation test. Table \ref{tab:table2} reports the classification performance by using the other five different learning strategies: the RNN, Deep Belief Networks (DBN), Convolutional Neural Network (CNN), SAE and Support Vector Machine with RBF kernel (SVMrbf). LSTM-RNN shows better accuracy and precision in all of the methods.
\begin{figure}[!t]
\setlength{\abovecaptionskip}{-0.55 cm}
\setlength{\belowcaptionskip}{-0.2 cm}
\centering
\vspace*{-0.8 cm}
\centerline{\includegraphics[width=0.75\linewidth]{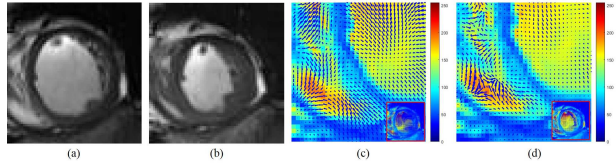}}
\caption{\footnotesize{A pair of frames at the beginning of systole (a) and at the end of systole (b) were first displayed, followed by the visual results of our deep optical flow (c) and Horn and Schunck(HS) optical flow (d) at pixel precision.}}
\label{fig:Figure4}
\vspace*{-0.4 cm}
\end{figure}
\vspace*{-0.15 cm}
\begin{table}[!hp]
\footnotesize
\setlength{\abovecaptionskip}{-0 cm}
\setlength{\belowcaptionskip}{-0 cm}
\vspace*{-0.35 cm}
  \centering
  \caption{LSTM-RNN works best in comparison with other models.}
    \begin{tabular}{ccccccc}
    \hline\noalign{\smallskip}
          & SVMrbf & SAE-3 & DBN-3 & CNN   & RNN   & LSTM-RNN \\
    \hline\noalign{\smallskip}
    Accuracy & 80.9\% & 83.5\%\% & 84.9\% & 83.7\% & 88.4\% & \textbf{94.3\%} \\
    Precision & 74.2\% & 75.5\% & 75.1\% & 76.5\% & 84.8\% & \textbf{91.3\%} \\
    \hline\noalign{\smallskip}
    \end{tabular}%
  \label{tab:table2}%
   \vspace*{-0.55 cm}
\end{table}%
\\\textbf{Performance of the optical flow.} The purpose of the optical flow is to capture the global motion features. To evaluate the performance of our optical flow algorithm with a deep architecture, we used the average angular error (AAE) to evaluate our deep optical flow and other optical flow approaches. The other optical flow methods, including the Horn and Schunck method, pyramid Horn and Schunck method, intensity-based optical flow method, and phase-based optical flow method, can be found in \cite{fortun2015optical}. The comparison results are shown in Table 3, and visual examples are illustrated in Fig. \ref{fig:Figure4}.
\begin{table}[!hp]
\footnotesize
\setlength{\abovecaptionskip}{-0 cm}
\setlength{\belowcaptionskip}{-0 cm}
\vspace*{-0.55 cm}
 \centering
  \caption{Deep optical flow (OF) can work better in comparison to other optical flow techniques in capturing global motion features.}
    \begin{tabular}{cccccc}
    \hline\noalign{\smallskip}
     &\scriptsize{Horn and Schunck(HS)} & \scriptsize{Pyramid HS} & \scriptsize{Deep OF} & \scriptsize{Intensity-based OF} & \scriptsize{Phase-based OF} \\
    \hline\noalign{\smallskip}
    OF density & 100\% & 100\% & \textbf{100\%} & 55\% & 13\% \\
    AAE & 12.6$^\circ$$\pm$9.2$^\circ$ & 7.4$^\circ$$\pm$3.4$^\circ$ & \textbf{5.7$^\circ$$\pm$2.3$^\circ$} & 5.7$^\circ$$\pm$4.1$^\circ$ & 5.5$^\circ$$\pm$3.9$^\circ$ \\
    \hline\noalign{\smallskip}
    \end{tabular}%
  \label{tab:table3}%
  \vspace*{-0.6 cm}
\end{table}%
\vspace*{-0.5 cm}\section{Conclusions}
\vspace*{-0.15 cm}We have, for the first time, developed and presented an end-to-end deep-learning framework for the detection of infarction areas at the pixel level from CMR sequences. Our experimental analysis was conducted on 114 subjects, and it yielded an overall classification accuracy of 94.35\% at the pixel level. All of these results demonstrate that our proposed method can aid in the clinical diagnosis of MI assessments.
\vspace*{-0.4 cm}
\section*{Acknowledgments}
\vspace*{-0.25 cm}This work was supported in part by the Shenzhen Research and Innovation Funding (JCYJ20151030151431727, SGLH20150213143207911), the National Key Research and Development Program of China (2016YFC1300302, 2016YFC13017
\\
00), the CAS President’s International Fellowship for Visiting
Scientists (2017VTA0011), the National Natural Science Foundation of China (No.61673020), the Provincial Natural Science Research Program of Higher Education Institutions of Anhui province (KJ2016A016) and the Anhui Provincial Natural Science Foundation (1708085QF143).
\vspace*{-0.4 cm}
\bibliographystyle{splncs03}

\vspace*{-0.9 cm}
\end{document}